\documentclass[twoside]{article}

\usepackage[preprint]{aistats2026}

\usepackage{myaistats26}
\usepackage{mynotation}
\usepackage{local-notation}

\usepackage[header]{appendix}
\usepackage{titletoc}

\begin{document}

\twocolumn[

\aistatstitle{On Measuring Localization of Shortcuts in Deep Networks}

\aistatsauthor{ Nikita Tsoy \And Nikola Konstantinov }

\aistatsaddress{
    INSAIT,\\Sofia University ``St. Kliment Ohridski'',\\Sofia, Bulgaria \And
    INSAIT,\\Sofia University ``St. Kliment Ohridski'',\\Sofia, Bulgaria } ]

\begin{abstract}
    Shortcuts, spurious rules that perform well during training but fail to
    generalize, present a major challenge to the reliability of deep networks
    \citep{g20s}. However, the impact of shortcuts on feature representations
    remains understudied, obstructing the design of principled
    shortcut-mitigation methods. To overcome this limitation, we investigate
    the layer-wise localization of shortcuts in deep models. Our novel
    experiment design quantifies the layer-wise contribution to accuracy
    degradation caused by a shortcut-inducing skew by counterfactual training
    on clean and skewed datasets. We employ our design to study shortcuts on vision tasks:
    CIFAR-10, Waterbirds, and CelebA, across VGG, ResNet, DeiT, and
    ConvNeXt architectures. We find that shortcut learning is not localized in
    specific layers but distributed throughout the network. Different network
    parts play different roles in this process: shallow layers predominantly
    encode spurious features, while deeper layers predominantly forget core
    features that are predictive on clean data. We also analyze the differences
    in localization and describe its principal axes of variation. Finally,
    our analysis of layer-wise shortcut-mitigation strategies suggests the
    hardness of designing general methods, supporting dataset- and
    architecture-specific approaches instead.
\end{abstract}

\section{INTRODUCTION}

\emph{Shortcuts}, spurious rules that hold within training distributions but
fail to generalize to real-world scenarios, present a major challenge to
the reliability of deep networks. Despite their significance, the mechanisms
underlying shortcut learning
remain poorly understood.
While shortcuts arise from a statistical phenomenon
of spurious correlations \citep{a20i}, it remains unclear how and which
correlations are captured during training \citep{h20w}.

A crucial yet underexplored dimension of shortcut learning is the
hierarchical nature of deep networks. Since different layers correspond to
distinct levels of abstraction and feature complexity \citep{s14d}, shortcuts
likely manifest differently across layers.
Quantifying the impact of this
phenomenon on accuracy could inform the design of layer-specific interventions
for shortcut mitigation \citep[e.g.,][]{l23s}. However, existing research falls
short in this regard---either focusing solely on overall model accuracy without
layer-specific effects \citep{s22w} or analyzing feature
representations without explicitly connecting these findings to test
performance \citep{h20w,i21s}.

\paragraph{Contributions}

To bridge this gap, we develop a method for quantifying the layer-wise effects
of shortcuts on model accuracy. Our approach measures \emph{each
layer's contribution to shortcut learning}, expressed as accuracy degradation
on clean test data, by analyzing \emph{counterfactual changes} in network
behavior when trained on skewed and clean data. We decompose shortcut learning
into two fundamental processes: spurious feature promotion and core feature
degradation. And, we analyze them using two metrics: spurious feature encoding
and core feature forgetting.

We apply our methodology in the context of vision models. In particular, we
study a watermark skew on CIFAR-10, background skew on Waterbirds, and sampling
skew on CelebA, across the VGG-11, ResNet-18, DeiT-Ti, and ConvNeXt-T
architectures. Our findings reveal that shortcut learning is not localized in
specific layers, but is instead distributed throughout the whole network.
Different layers play different roles in shortcut learning: shallow
layers mostly contribute to spurious feature encoding, while deep layers mostly
contribute to core feature forgetting. Dataset and model factors explain
$87.0\%$ of variance in encoding localization, while data skew frequency and
optimizer factors explain $62.3\%$ of variance in forgetting localization in
fine-tuned models. Finally, we find that our metrics predict the success of
some layer-wise interventions.

From a practical standpoint, localization varies a
lot across different datasets and architectures. Additionally, our localization
metrics predict the success of some layer-wise specific interventions (e.g.,
layer freezing). Together, these findings suggest that layer-wise vision shortcut mitigation
strategies should be dataset- and architecture-specific.

\section{RELATED WORK}

\paragraph{Impact of Shortcuts on Feature Representations}
Several studies examine the impact of shortcuts on representations
\citep{h20w,i21s,s22w}. They analyze how shortcuts are encoded
in layers through linear probing accuracy \citep{h20w}, mutual information and
read-out module accuracy \citep{i21s}, or validation accuracy on
feature-labeled datasets \citep{s22w}. While these approaches
provide valuable insights, they do not explicitly attribute accuracy
degradation to specific layers. In contrast, our work directly quantifies
\emph{layer-wise contributions to accuracy degradation}, offering a more direct
assessment of each layer's role in shortcuts.

\paragraph{Mechanisms of Shortcut Formation}
Our work contributes to the literature on shortcut learning mechanisms
\citep{s20t,s20a,n21u,c23w,p23d,w23w,t24s}. Our analysis suggests
that feature forgetting plays a key role in shortcuts, supporting prior
hypotheses that simplicity bias \citep[e.g.,][]{s20t} or excessive
regularization \citep{s20a} are important for shortcut formation.
In contrast to these works, we measure the contributions to shortcut learning
by different parts of the network, allowing for a fine-grained quantitative
understanding of shortcuts.

\paragraph{Quantification of Layers' Importance}
Similarly to us, \citet{z22a,m23c,h23t} investigate feature
representations in deep models and assess the importance of each layer for
specific model properties. \citet{z22a} measure the importance of each layer of
a deep network for classification accuracy by injecting noise in the network
weights. \citet{m23c} analyze the memorization behavior of different layers by
introducing label noise. \citet{h23t} analyze learned feature representations
of deep models and show how some of their properties help generalization. While
these studies provide valuable insights into feature learning, they do not
study shortcut learning. Thus, these approaches are not directly comparable to
our methodology.

\paragraph{Layer-Wise Fine-Tuning Analysis}
Our work is related to the literature on layer-wise adaptation of deep
models to distribution shifts \citep[e.g.,][]{k22f,l23s,t23a,k23l}.
Our findings help to reason about fine-tuning strategies
for shortcut mitigation. We refer to our conclusion section for further
discussion.

\section{METHODOLOGY}

This section outlines our method for measuring \emph{layer-wise contributions}
to shortcut learning. Our approach introduces controlled
\emph{shortcut-inducing skews} into the training process, such as replacing the
background of an image with a class-correlated one, like in Waterbirds
\citep{s20d}. This approach allows us to assess each layer's role
\emph{counterfactually}, by controlling all factors of training except for the
data itself. We train multiple networks on the same task, exposing different
blocks of layers to skewed or clean (skew-free) data and evaluate
these networks on a clean test dataset to quantify each block's contribution.

\subsection{Illustrative Example}
\label{sec:example}

We start with an architecture $h = (c,f)$ consisting of a classifier $c$ and
feature extractor $f$: $h(\cdot) = c(\cdot) \circ f(\cdot)$.
Consider training this architecture on two datasets: a clean
\textcolor{blue}{$D^c$} and a skewed \textcolor{red}{$D^s$} ones, resulting in
two models $\textcolor{blue}{h^c} = (\textcolor{blue}{c^c},
\textcolor{blue}{f^c})$ and $\textcolor{red}{h^s} = (\textcolor{red}{c^s},
\textcolor{red}{f^s})$, respectively. Due to shortcut learning, we expect to
see an increase in test error rate on clean data, $\err(\textcolor{red}{h^s}) -
\err(\textcolor{blue}{h^c})$.

We aim to quantify the contributions of the classifier $c$ and feature
extractor $f$ to the increase in test error rate $\err(\textcolor{red}{h^s}) -
\err(\textcolor{blue}{h^c})$. Specifically, we are interested in how spurious
(that emerge due to the skew) and core features (that are predictive on clean
data) are encoded and used. Let $\textcolor{blue}{c^{c, \textcolor{red}{f^s}}}$
denote a classifier retrained on clean data using the skewed extractor
$\textcolor{red}{f^s}$ and $\textcolor{red}{c^{s, \textcolor{blue}{f^c}}}$
denote a classifier retrained on skewed data using the clean extractor
$\textcolor{blue}{f^c}$ (see details in \cref{sec:training}).  We consider the
following two decompositions:
\begin{equation}
    \begin{aligned}
        \err&(\textcolor{red}{h^s}) - \err(\textcolor{blue}{h^c})\\
        = & \err(\textcolor{red}{c^s}, \textcolor{red}{f^s}) -
        \err(\textcolor{red}{c^{s, \textcolor{blue}{f^c}}},
        \textcolor{blue}{f^c})
        + \err(\textcolor{red}{c^{s, \textcolor{blue}{f^c}}},
        \textcolor{blue}{f^c}) -
        \err(\textcolor{blue}{c^c}, \textcolor{blue}{f^c})\\
        = & \err(\textcolor{red}{c^s}, \textcolor{red}{f^s}) -
        \err(\textcolor{blue}{c^{c, \textcolor{red}{f^s}}},
        \textcolor{red}{f^s})
        + \err(\textcolor{blue}{c^{c, \textcolor{red}{f^s}}},
        \textcolor{red}{f^s}) -
        \err(\textcolor{blue}{c^c}, \textcolor{blue}{f^c}).
    \end{aligned}
\end{equation}
We interpret the first decomposition in the following manner.
In the term $\err(\textcolor{red}{c^s}, \textcolor{red}{f^s}) -
\err(\textcolor{red}{c^{s, \textcolor{blue}{f^c}}}, \textcolor{blue}{f^c})$,
both classifiers are trained on the skewed data. Hence, this term isolates the
effect of the extractor on the propensity of the classifier to rely on spurious
correlations. Thus, this term measures the increase in \emph{spurious feature
encoding} by the skewed feature extractor compared to the clean one. In the
term $\err(\textcolor{red}{c^{s, \textcolor{blue}{f^c}}},
\textcolor{blue}{f^c}) - \err(\textcolor{blue}{c^c}, \textcolor{blue}{f^c})$,
both models use the same extractor. Hence, this term isolates the effect of
training data on the reliance of the classifier on core features. Thus, this
term measures the \emph{core feature underutilization} caused by the skew in
the retrained classifier.

Similarly,
$\err(\textcolor{red}{c^s}, \textcolor{red}{f^s}) -
\err(\textcolor{blue}{c^{c, \textcolor{red}{f^s}}}, \textcolor{red}{f^s})$ measures
\emph{spurious feature amplification} in the skewed classifier and
$\err(\textcolor{blue}{c^{c, \textcolor{red}{f^s}}}, \textcolor{red}{f^s}) -
\err(\textcolor{blue}{c^c}, \textcolor{blue}{f^c})$ corresponds to \emph{core
feature forgetting} by the skewed feature extractor.\footnote{While we expect
all considered differences to be positive,
it is generally not guaranteed. For example, some layers might be responsible
for filtering out a shortcut rule because it does not have perfect predictive
power.}

\paragraph{Shortcut Learning Metrics}
To measure how spurious and core features are localized and used, we use two
metrics: \emph{spurious feature encoding} and \emph{core feature forgetting}.
They determine how much the extractor ``incentivizes'' the classifier
to rely on spurious features or ``disincentives'' the classifier to rely clean
features, respectively. In our work, we aim to localize blocks' contributions
to these phenomena to better understand the mechanisms of shortcut learning.
Note also that the other metrics: \emph{spurious feature amplification} and
\emph{core feature underutilization}, complement the chosen ones, making
their localization symmetrical.

\paragraph{Counterfactual Training Algorithm}
A crucial aspect for objectively measuring these metrics is the design of an
appropriate classifier retraining scheme.
Retraining needs to maintain consistency between the learning
mechanisms in the original and new classifier retraining, i.e., it should
\textit{control for all factors apart from the training data itself}, to
avoid biases in localization measures.
At the same time, due to overparametrization \citep{b22i} and implicit biases
\citep[e.g.,][]{n21u}, deep learning is highly sensitive to even
small changes in training procedure, making this task non-trivial.


\subsection{Shortcut Learning Metrics}
\label{subsec:metrics}

In the general case, we consider a feed-forward architecture consisting of $m$
blocks of layers
\begin{equation}
    f(\theta, \cdot) = f_{m-1}(\theta_{m-1}, \cdot) \circ f_{m-2}(\theta_{m-2},
    \cdot) \circ f_0(\theta_0, \cdot).
\end{equation}
Let $\theta_A$ represent the weights of blocks $i \in A$, where $A \subseteq
[m]$ and $[m] \defeq \{0, \dots, m-1\}$. Define $i \mathbin{:} j \defeq \{i,
\dots, j-1\}$, and let $\err(\theta)$ be the error rate of this architecture
with weights $\theta$ on the clean test dataset. Consider two networks trained
on clean and skewed data, resulting in weights $\theta^c$ and $\theta^s$,
respectively. We interpret the increase in error rate $\err(\theta^s) -
\err(\theta^c)$ as a measure of shortcut learning. Our goal is to measure the
contributions of individual blocks to this increase.

We investigate what would happen if some intervention subset of blocks $A$ of
the clean (skewed) model were \emph{counterfactually trained} on skewed (clean)
data. Let $\theta^{c, A}$ be a model that shares blocks $[m] \setminus A$ with
the clean model, but whose subset of blocks $A$ was counterfactually trained on
the skewed data (with $\theta^{s, A}$ defined analogously). As previously, we
get decompositions
\[
    \err(\theta^s) - \err(\theta^c) = \enc_{[m] \setminus A} + \uut_{A} =
    \amp_{A} + \fgt_{[m] \setminus A},
\]
where
$\enc_{[m] \setminus A} \defeq \err(\theta^s) - \err(\theta^{c, A})$ is the
contributions of blocks $[m] \setminus A$ to \emph{spurious features encoding},
$\uut_{A} \defeq \err(\theta^{c, A}) - \err(\theta^c)$ is the contributions of
blocks $A$ to \emph{core features underutilization},
$\fgt_{[m] \setminus A} \defeq \err(\theta^{s, A}) - \err(\theta^c)$ is the
contributions of blocks $[m] \setminus A$ to \emph{core features forgetting},
and $\amp_{A} \defeq \err(\theta^s) - \err(\theta^{s, A})$ is the contributions
of blocks $A$ to \emph{spurious features amplification}.
We often consider \emph{relative contributions} normalized with $\err(\theta^s)
- \err(\theta^c)$ to allow fair comparison across datasets and architectures
(see \cref{sec:exper-details} for details).

Again, we can see that shortcut learning encompasses two processes: one related
to core feature degradation (expressed as core feature forgetting or
underutilization) and one related to spurious feature promotion (expressed as
spurious feature amplification or encoding). Our framework allows us to
quantify how different network parts contribute to these processes.
Specifically, we focus on the spurious feature encoding and core feature
forgetting metrics, with results on the localization of the other metrics being
symmetrical.

\subsection{Counterfactual Training Algorithm}
\label{sec:training}

As previously argued, a key challenge in our approach is the design of a
\emph{counterfactual training} procedure that preserves learning mechanisms
across all models. There are several critical factors: ensuring all
corresponding blocks have \emph{equal exposure to training data}, maintaining
\emph{consistent optimizer configurations and hyperparameters} \citep[as some
hyperparameters, such as learning rate, significantly impact
features,][]{l19t,l20l}, and allowing blocks to \emph{progressively adapt} to
intermediate features since progressive adaptation invokes different
learning mechanisms compared to static post-training \citep{a19w,p24p,a22t}.

We solve this challenge through a simultaneous training procedure described in
\cref{alg:sim-train} (for brievity, we only consider training $\theta^{c, A}$
with stochastic gradient decent (SGD), but extensions to training $\theta^{s,
A}$ and other optimizers are straightforward). Here, the loss of a network with
weights $\theta$ on a data batch $B$ is denoted by $L(\theta, B)$. This
procedure trains the anchor clean network $\theta^c$ and counterfactually
intervened network $\theta^{c, A}$ in parallel. In each round, we sample a
clean data batch $B_t$ and skewed data batch $B'_t = g(B_t)$ to update the
models. Since blocks $[m] \setminus A$ are shared, we essentially only update
$\theta^{c, A}_A$ using skewed data during the backward pass through the
network $\theta^{c, A}$.

\begin{algorithm}[htb]
    \caption{Simultaneous training of networks}
    \label{alg:sim-train}
    \begin{algorithmic}
        \STATE Initialize $\theta^c_0$ --- anchor network weights
        \STATE Initialize $\theta^{c, A}_0 = \theta^c_0$ --- intervened network
        weights
        \FOR{$t=1$ {\bfseries to} $T$}
            \STATE Sample clean $B_t$ and skewed $B'_t = g(B_t)$ batches
            \STATE Update $\theta^{c, A}_t = \theta^{c, A}_{t-1} - \eta_t
            \nabla L(\theta^{c, A}_{t-1}, B'_t)$
            \STATE Update $\theta^c_t = \theta^c_{t-1} - \eta_t \nabla
            L(\theta^c_{t-1}, B_t)$
            \STATE Synchronize shared blocks $\theta^{c, A}_{t, [m] \setminus
            A} = \theta^c_{t, [m] \setminus A}$
        \ENDFOR
    \end{algorithmic}
\end{algorithm}

Throughout this process, each skewed counterfactually trained block
progressively adapts to the neighboring clean blocks to classify skewed data.
By design, all blocks receive the same exposure to training data, with the only
difference being the presence or the absence of a skew. The training algorithm
remains consistent across all blocks, controlling for the optimizer's implicit
biases and satisfying our desiderata.

\section{EXPERIMENT DETAILS}
\label{sec:exper-details}

\begin{figure}[htb]
    \centering
    \includegraphics[width=0.4\linewidth]{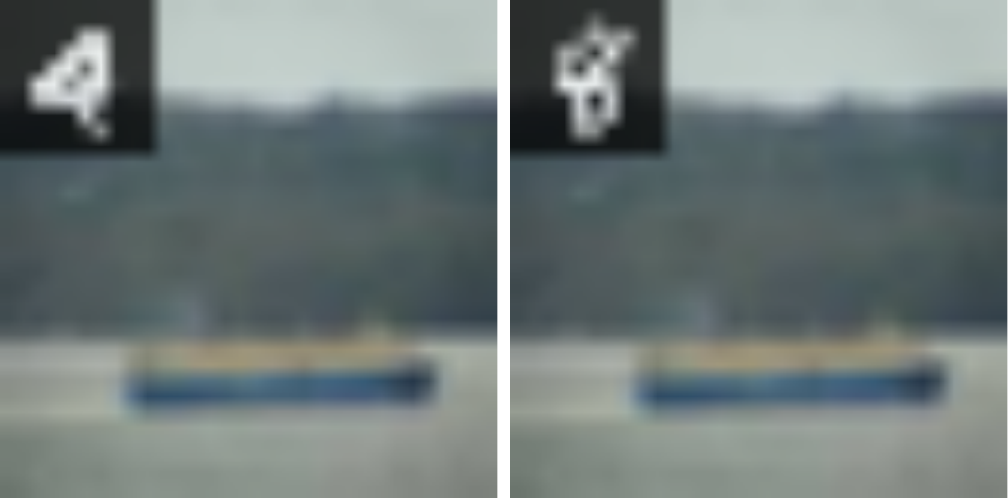}
    \caption{Clean (left) and skewed (right) CIFAR-10 image of class 8 with
    MNIST watermark}
    \label{fig:watermark-example}
\end{figure}

\paragraph{Datasets and Skews}
We consider three datasets with distinct skews: CIFAR-10 \citep{k09l} with
\emph{watermark skew}, Waterbirds \citep{s20d} with \emph{background skew}, and
CelebA \citep{l15d} with \emph{group sampling skew} \citep{s20d}. For the
watermark skew (see \cref{fig:watermark-example}), we blend the upper-left
corner of CIFAR images with class-correlated MNIST \citep{l98g} digits,
encouraging the network to rely on the simple MNIST watermark. For the
background skew, following \citet{s20d}, we place bird images on
class-correlated backgrounds, incentivizing the reliance on background cues.
For the sampling skew, we sample the skewed dataset introducing a correlation
between gender and hair color, encouraging demographic shortcuts.

We generate skews using the following procedure. First, we create a clean
dataset where the considered skew is not predictive. For CIFAR and
Waterbirds, we simply match each image with a random image of MNIST digit or
background, respectively. For CelebA, we remove some images of
blond females, non-blond males, and non-blond females to balance
groups.\footnote{The final number of blond and non-blond images equals to the
number of blond females and non-blond males in the original dataset} Then, we create a
\emph{fully skewed dataset} where the considered skew is perfectly predictive
of the label. For CIFAR and Waterbirds, we replace a previously matched random
image with an image corresponding to the image label if the class of random
image does not already correspond to the image label. For CelebA, we replace
the images of blonde males with blonde females and the images of non-blonde
females with non-blonde males, making spurious correlation perfectly
predictive. Finally, we create a skewed dataset, where each clean image is
replaced with a corresponding fully skewed image with a certain frequency. We
consider two frequencies: \emph{common} and \emph{rare} ($\nicefrac{127}{128}$
and $\nicefrac{15}{16}$, respectively). For CIFAR, we use watermarks of size
$10 \times 10$ and two blending strengths \emph{strong} and \emph{weak}, which
equal to $\nicefrac{3}{4}$ and $\nicefrac{1}{4}$, respectively.

\paragraph{Models and Optimizers} We consider four architectures: VGG-11
\citep{s15v}, a typical convolutional neural network (CNN); ResNet-18
\citep{h16d}, a CNN with residual connections; DeiT-Ti \citep{t21t}, a vision
transformer (ViT); and ConvNeXt-T \citep{l22a}, a modernized CNN. Since all
tasks have few classes, we use global average pooling instead of dense
classification layers in VGG-11. We decompose each architecture into 6 blocks.
The first block always corresponds to the initial convolutional layer, while
the last block corresponds to the final linear layer. For the rest, we do the
following. For VGG, we use max-pool layers as block boundaries. For ResNet and
ConvNeXt, we use convolutional layers with stride 2 as boundaries \citep[see
Figure 3 in][]{h16d}. For DeiT, we divide the rest into four equal blocks. We
consider SGD \citep{r51a} and AdamW \citep{l18d} optimizers, and either train
from scratch or fine-tune from ImageNet \citep{d09i} initialization. (See
\cref{sec:train-details} for details.)

\paragraph{Experiment Scope} We conduct two types of experiments. First, using
intervention sets $A = [m] \setminus \{i\}$, we assess whether shortcuts could
be localized in a single block by testing whether a
\emph{single block's contribution} to could fully explain them, i.e., $\exists
i : \frac{\fgt_{i}}{\err(\theta^s) - \err(\theta^c)} \approx 1
\lor \frac{\enc_{i}}{\err(\theta^s) - \err(\theta^c)} \approx 1$. Second, to
quantify layer-wise contributions, using intervention sets $A = i
\mathbin{:} m$, we measure the \emph{rate of increase in relative contributions
of initial blocks'} to forgetting and encoding,
\[
    \frac{\fgt_{0:i+1} - \fgt_{0:i}}{\err(\theta^s) - \err(\theta^c)}
    \quad \text{and} \quad
    \frac{\enc_{0:i+1} - \enc_{0:i}}{\err(\theta^s) - \err(\theta^c)}.
\]

\begin{table*}[htb]
    \centering
    \caption{Average clean test error rates (and their standard errors in
    parenthesis) of clean and skewed models fine-tuned with AdamW for rare
    (top part) and common (bottom part) skews}
    \label{tab:err-rates}
    \begin{tabular}{l c c c c c c c c}
\toprule
& \multicolumn{2}{c}{CIFAR-10 (weak)} & \multicolumn{2}{c}{CIFAR-10 (strong)} & \multicolumn{2}{c}{Waterbirds} & \multicolumn{2}{c}{CelebA} \\
\cmidrule(lr){2-3} \cmidrule(lr){4-5} \cmidrule(lr){6-7} \cmidrule(lr){8-9}
Model & Clean & Skewed & Clean & Skewed & Clean & Skewed & Clean & Skewed \\
\midrule
DeiT-Ti & $3.2\%$ & $5.2\%$ & $3.4\%$ & $8.6\%$ & $2.0\%$ & $9.4\%$ & $5.8\%$ & $11.3\%$ \\
& $(0.1\%)$ & $(0.1\%)$ & $(0.1\%)$ & $(0.1\%)$ & $(0.1\%)$ & $(0.4\%)$ & $(0.1\%)$ & $(0.2\%)$ \\
ResNet-18 & $4.1\%$ & $7.2\%$ & $4.3\%$ & $11.5\%$ & $2.1\%$ & $8.5\%$ & $5.9\%$ & $11.3\%$ \\
& $(0.1\%)$ & $(0.1\%)$ & $(0.1\%)$ & $(0.2\%)$ & $(0.1\%)$ & $(0.3\%)$ & $(0.1\%)$ & $(0.3\%)$ \\
VGG-11 & $6.9\%$ & $17.1\%$ & $7.2\%$ & $26.3\%$ & $2.5\%$ & $10.7\%$ & $6.0\%$ & $11.0\%$ \\
& $(0.1\%)$ & $(0.3\%)$ & $(0.1\%)$ & $(0.2\%)$ & $(0.1\%)$ & $(0.2\%)$ & $(0.1\%)$ & $(0.3\%)$ \\
ConvNext-T & $1.8\%$ & $3.3\%$ & $1.9\%$ & $5.1\%$ & $0.8\%$ & $3.7\%$ & $5.8\%$ & $11.4\%$ \\
& $(0.1\%)$ & $(0.1\%)$ & $(0.1\%)$ & $(0.1\%)$ & $(0.1\%)$ & $(0.2\%)$ & $(0.1\%)$ & $(0.2\%)$ \\
\midrule
DeiT-Ti & $3.2\%$ & $7.8\%$ & $3.4\%$ & $17.5\%$ & $2.0\%$ & $19.9\%$ & $5.8\%$ & $20.4\%$ \\
& $(0.1\%)$ & $(0.2\%)$ & $(0.1\%)$ & $(0.3\%)$ & $(0.1\%)$ & $(0.6\%)$ & $(0.1\%)$ & $(0.8\%)$ \\
ResNet-18 & $4.1\%$ & $11.3\%$ & $4.3\%$ & $25.3\%$ & $2.1\%$ & $18.2\%$ & $5.9\%$ & $19.4\%$ \\
& $(0.1\%)$ & $(0.3\%)$ & $(0.1\%)$ & $(0.3\%)$ & $(0.1\%)$ & $(0.2\%)$ & $(0.1\%)$ & $(0.4\%)$ \\
VGG-11 & $6.9\%$ & $29.2\%$ & $7.2\%$ & $52.4\%$ & $2.5\%$ & $25.3\%$ & $6.0\%$ & $21.3\%$ \\
& $(0.1\%)$ & $(0.5\%)$ & $(0.1\%)$ & $(0.4\%)$ & $(0.1\%)$ & $(0.4\%)$ & $(0.1\%)$ & $(0.7\%)$ \\
ConvNext-T & $1.8\%$ & $5.4\%$ & $1.9\%$ & $11.6\%$ & $0.8\%$ & $10.9\%$ & $5.8\%$ & $20.8\%$ \\
& $(0.1\%)$ & $(0.1\%)$ & $(0.1\%)$ & $(0.2\%)$ & $(0.1\%)$ & $(0.5\%)$ & $(0.1\%)$ & $(0.7\%)$ \\
\bottomrule
    \end{tabular}
\end{table*}

\begin{table*}[htb]
    \centering
    \caption{Average relative contributions (and standard errors) of single
    blocks to encoding (top) and forgetting (bottom) for models fine-tuned with
    AdamW on CIFAR-10 (strong) with common skew}
    \label{tab:cifar-ind}
    \begin{tabular}{l c c c c c c}
\toprule
Model & Bl. $0$ & Bl. $1$ & Bl. $2$ & Bl. $3$ & Bl. $4$ & Bl. $5$ \\
\midrule
DeiT-Ti & $0.1\%$ & $33.5\%$ & $42.9\%$ & $48.8\%$ & $25.4\%$ & $0.1\%$ \\
& $(1.7\%)$ & $(2.2\%)$ & $(2.0\%)$ & $(0.9\%)$ & $(2.3\%)$ & $(1.7\%)$ \\
ResNet-18 & $2.4\%$ & $9.8\%$ & $16.9\%$ & $46.1\%$ & $66.0\%$ & $2.7\%$ \\
& $(1.2\%)$ & $(1.0\%)$ & $(1.4\%)$ & $(0.6\%)$ & $(0.5\%)$ & $(0.7\%)$ \\
VGG-11 & $-0.5\%$ & $-0.2\%$ & $12.0\%$ & $53.0\%$ & $39.8\%$ & $4.1\%$ \\
& $(0.3\%)$ & $(0.2\%)$ & $(0.8\%)$ & $(1.7\%)$ & $(2.6\%)$ & $(0.5\%)$ \\
ConvNext-T & $-1.9\%$ & $4.5\%$ & $13.2\%$ & $76.4\%$ & $44.1\%$ & $0.1\%$ \\
& $(3.5\%)$ & $(1.3\%)$ & $(3.0\%)$ & $(1.2\%)$ & $(0.7\%)$ & $(1.6\%)$ \\
\midrule
DeiT-Ti & $0.7\%$ & $1.0\%$ & $0.8\%$ & $0.4\%$ & $0.6\%$ & $-0.2\%$ \\
& $(0.5\%)$ & $(0.4\%)$ & $(0.6\%)$ & $(0.4\%)$ & $(1.0\%)$ & $(0.4\%)$ \\
ResNet-18 & $0.2\%$ & $0.6\%$ & $1.8\%$ & $2.7\%$ & $5.1\%$ & $0.3\%$ \\
& $(0.4\%)$ & $(0.3\%)$ & $(0.4\%)$ & $(0.6\%)$ & $(0.2\%)$ & $(0.2\%)$ \\
VGG-11 & $-0.1\%$ & $0.2\%$ & $0.8\%$ & $10.0\%$ & $9.7\%$ & $0.3\%$ \\
& $(0.3\%)$ & $(0.3\%)$ & $(0.3\%)$ & $(1.4\%)$ & $(1.7\%)$ & $(0.2\%)$ \\
ConvNext-T & $-0.4\%$ & $1.0\%$ & $0.8\%$ & $1.6\%$ & $1.0\%$ & $-0.6\%$ \\
& $(0.9\%)$ & $(1.0\%)$ & $(0.8\%)$ & $(0.7\%)$ & $(0.8\%)$ & $(0.6\%)$ \\
\bottomrule
    \end{tabular}
\end{table*}

\begin{table*}[htb]
    \centering
    \caption{Relative forgetting rate (and standard errors)
    of DeiT-Ti models fine-tuned on CIFAR-10 (strong)}
    \label{tab:forget-decomp}
    \begin{tabular}{l l c c c c c c}
\toprule
Frequency & Optimizer & Bl. $0$ & Bl. $1$ & Bl. $2$ & Bl. $3$ & Bl. $4$ & Bl. $5$ \\
\midrule
Rare & AdamW & $0.1\%$ & $3.2\%$ & $5.0\%$ & $8.4\%$ & $19.8\%$ & $63.7\%$ \\
& & $(2.2\%)$ & $(0.8\%)$ & $(1.3\%)$ & $(1.7\%)$ & $(2.0\%)$ & $(1.8\%)$ \\
& SGD & $0.3\%$ & $4.4\%$ & $4.0\%$ & $9.1\%$ & $11.0\%$ & $71.6\%$ \\
& & $(1.4\%)$ & $(1.0\%)$ & $(1.0\%)$ & $(0.5\%)$ & $(0.9\%)$ & $(1.2\%)$ \\
\midrule
Common & AdamW & $0.6\%$ & $1.1\%$ & $1.9\%$ & $5.2\%$ & $11.0\%$ & $80.6\%$ \\
& & $(0.5\%)$ & $(0.5\%)$ & $(0.4\%)$ & $(0.2\%)$ & $(0.5\%)$ & $(0.8\%)$ \\
& SGD & $0.2\%$ & $1.9\%$ & $1.9\%$ & $5.1\%$ & $6.7\%$ & $84.5\%$ \\
& & $(0.2\%)$ & $(0.4\%)$ & $(0.4\%)$ & $(0.6\%)$ & $(0.6\%)$ & $(0.9\%)$ \\
\bottomrule
    \end{tabular}
\end{table*}

\section{RESULTS}
\label{sec:results}

This section presents the results of our experiments for fine-tuned models (see
training from scratch results in \cref{sec:scratch}). First, we compute the
localization metrics for individual and initial blocks. Then, we analyze
which factors contribute the most to the localization in the initial blocks.
Finally, we investigate whether our localization metrics are predictive of the
success of layer-wise training interventions.

To understand the extent of shortcut learning, \cref{tab:err-rates} presents the
clean test dataset error rates achieved by the models fine-tuned with AdamW on
clean and skewed data for our datasets (SGD behavior is similar). We repeat each experiment 5 times to average over training noise and report the
averaged values and their standard errors with the Bessel's correction.

\subsection{Localization of Encoding and Forgetting}

\paragraph{Localization in a Single Block}
\cref{tab:cifar-ind} presents the relative individual contributions of
blocks for the models fine-tuned with AdamW on the CIFAR-10 (strong) dataset
with common skew. First, no single block achieves $100\%$ relative contribution
to encoding or forgetting (according to the $5\%$ critical value for one sided
t-test). Moreover, there is no individual block whose contribution is much
larger than the contributions of other blocks. Second, the
sum of individual contributions generally either significantly (according to
the $5\%$ critical value for the two-sided t-test) exceeds $100\%$ (for
encoding, with an exception of VGG-11 model) or does not reach $100\%$ (for
forgetting), suggesting that simply analyzing individual block contributions
without interactions between them generally does not lead to accurate analysis.

Our findings suggest that shortcut learning is not concentrated in any single
block. The interactions between layers seem crucial for shortcut emergence,
suggesting that shortcut learning cannot be easily decomposed into a sum of
individual layer contributions.
Similar patterns emerge across all experimental settings (see additional results in \cref{sec:add-results}).
For single-block localization and fine-tuning, we discovered a rare possibility
of model divergence (when the model achieves a higher error rate than the
skewed model on clean data and a higher error rate than the clean model on
skewed data). Divergences occurred in only 7 (out of 3840) intervened models
(and only in the single block setting). Such models were excluded from the
analysis without impacting our conclusions.

\paragraph{Localization in the Initial Blocks}
To account for layer interactions, we examined intervention sets $A = i
\mathbin{:} m$. \cref{tab:init-adamw} presents the results for models
fine-tuned with AdamW on common skews. To make the results comparable, we
report the rate of increase in relative contributions. As expected, encoding
and forgetting generally increase with the number of layers involved.
All architectures first encode spurious features and subsequently
forget core features. The last layer plays a major role in feature
forgetting, while the first layer has minimal contributions.\footnote{For the
training from scratch, the first layer starts to engage in spurious
feature encoding, \cref{tab:init-adamw-s} in Appendix.}

These results indicate that shortcut learning indeed consists of two processes
that occur in different layers. First, the network encodes spurious features;
then, due to predictive spurious features, it progressively forgets core
features, leading to a shortcut classification rule.

\subsection{Differences in Relative Contributions}


\paragraph{Main Explanatory Factors}
\cref{tab:var-decomp} reports the fraction of the total variance of the
increase rate of relative encoding and forgetting explained by different
factors. Dataset and model architecture are the most predictive factors for
encoding localization, while skew frequency and optimizer choice are the most
predictive for forgetting localization. Together, dataset and model factors
explain $83.8\%$ of variance in encoding localization, while skew frequency and
optimizer factors explain $57.0\%$ of variance in forgetting localization.

\begin{table}[htb]
    \centering
    \caption{Variance explained in relative encoding (top) and forgetting
    (bottom) rate by different factors}
    \label{tab:var-decomp}
    \begin{tabular}{c c c c}
\toprule
Dataset & Skew freq. & Model & Optimizer \\
\midrule
$45.7\%$ & $0.4\%$ & $26.3\%$ & $3.3\%$ \\
$14.7\%$ & $21.0\%$ & $9.1\%$ & $33.5\%$ \\
\bottomrule
\end{tabular}
\end{table}

These findings suggest that encoding localization is primarily driven by the
skew's semantic properties: the dataset is directly related to it, while
the architecture determines the abstraction levels of layers. At the same
time, forgetting appears to be primarily influenced by the skew's predictive
power (through the dataset and skew frequency) and the implicit biases of
optimizer. We further explore these factors below.

\paragraph{Differences in Encoding}
\cref{tab:init-adamw} (top part) shows the increase rate in relative
encoding for models fine-tuned with AdamW on common skews. Watermark skew tends
to be encoded earlier, while sampling skew is typically encoded
in later layers. Additionally, CNN architectures generally encode spurious
features in the latter layers compared to ViT architectures.

\paragraph{Differences in Forgetting}
\cref{tab:forget-decomp} presents the increase rate in relative forgetting for
DeiT-Ti models fine-tuned on the CIFAR-10 (strong) dataset. SGD prioritizes
forgetting in the last layer, whereas AdamW primary distributes it between the
last layer and the penultimate block. Simultaneously, forgetting due to common
skews is relatively more concentrated in the last layer compared to one induced
by rare skews.

\subsection{Localization-Guided Interventions}

This section explores whether our localization metrics can predict the success
of shortcut mitigation strategies. We retrained DeiT-Ti, ResNet-18, and VGG-11
with AdamW on skewed data, using common skews in Waterbirds, CelebA, and
CIFAR-10 (strong). We consider four retraining interventions, where we modify the hyperparameters of the
optimizer layer-wise: 1--2) increasing or
decreasing LR (learning rate) by a factor of 3, and 3--4) increasing or
decreasing WD (weight decay) by a factor of 10. We applied these interventions
to individual blocks and the groups of two consecutive blocks. Then, we
regressed the extent of shortcut mitigation (the difference in test accuracy on
clean data between skewed retrained and non-intervened models normalized
against the same difference between non-intervened skewed and clean models), on
encoding and forgetting metrics (from \cref{tab:init-adamw}), their interaction
(i.e., a product of these metrics), their squares, and dummies for the first
and last layer, and double layer intervention. Additionally, we conduct an
experiment where we train the last layer and then the whole network for a short
time, and then freeze all blocks except one during fine-tuning.

\cref{tab:regression} presents the results for the relevancy of our metrics
(expressed through the F-statistic \citep{g03e} for the joint significance of the five metric-related coefficients) and overall predictive power of the regression
(expressed through R${}^2$).\footnote{see \cref{tab:full-regression} in Appendix for
regression coefficients} Our localization metrics are predictive of success
for LR and freezing interventions (on a 5\% significance level).
This finding suggests that our localization metrics capture relevant
information and could inform shortcut mitigations.

\begin{table}[htb]
    \centering
    \caption{Predictive power of localization metrics}
    \label{tab:regression}
    \begin{tabular}{l c c c c c}
\toprule
& LR$\uparrow$ & LR$\downarrow$ & WD$\uparrow$ & WD$\downarrow$ & Freeze \\
\midrule
F-Stat & $5.04$ & $8.76$ & $0.66$ & $0.44$ & $4.07$ \\
R${}^2$ & $0.36$ & $0.54$ & $0.07$ & $0.05$ & $0.32$ \\
N & $99$ & $99$ & $99$ & $99$ & $54$ \\
\bottomrule
    \end{tabular}
\end{table}

\begin{table*}[!tb]
    \centering
    \caption{Average increase rate in relative contributions (and standard
    errors) of the initial blocks to encoding (top part) and forgetting (bottom
    part) for models fine-tuned with AdamW on common skews}
    \label{tab:init-adamw}
    \begin{tabular}{l l c c c c c c}
\toprule
Dataset & Model & Bl. $0$ & Bl. $1$ & Bl. $2$ & Bl. $3$ & Bl. $4$ & Bl. $5$ \\
\midrule
CIFAR-10 & DeiT-Ti & $1.1\%$ & $40.3\%$ & $39.5\%$ & $15.1\%$ & $4.3\%$ & $0.1\%$ \\
(strong) & & $(3.1\%)$ & $(2.0\%)$ & $(1.8\%)$ & $(0.7\%)$ & $(0.4\%)$ & $(0.1\%)$ \\
& ResNet-18 & $2.4\%$ & $10.5\%$ & $26.1\%$ & $45.6\%$ & $15.7\%$ & $0.1\%$ \\
& & $(1.2\%)$ & $(1.0\%)$ & $(1.4\%)$ & $(0.8\%)$ & $(0.6\%)$ & $(0.1\%)$ \\
& VGG-11 & $-0.5\%$ & $1.5\%$ & $17.8\%$ & $72.3\%$ & $9.1\%$ & $0.0\%$ \\
& & $(0.3\%)$ & $(0.2\%)$ & $(0.3\%)$ & $(0.2\%)$ & $(0.3\%)$ & $(0.1\%)$ \\
& ConvNext-T & $-1.8\%$ & $6.7\%$ & $24.6\%$ & $67.6\%$ & $3.3\%$ & $-0.1\%$ \\
& & $(3.8\%)$ & $(5.3\%)$ & $(2.1\%)$ & $(1.9\%)$ & $(0.2\%)$ & $(0.2\%)$ \\
\midrule
Waterbirds & DeiT-Ti & $1.2\%$ & $19.1\%$ & $30.3\%$ & $24.6\%$ & $25.0\%$ & $0.0\%$ \\
& & $(2.4\%)$ & $(2.5\%)$ & $(2.8\%)$ & $(1.8\%)$ & $(1.1\%)$ & $(0.1\%)$ \\
& ResNet-18 & $6.7\%$ & $11.7\%$ & $18.9\%$ & $27.2\%$ & $35.4\%$ & $0.4\%$ \\
& & $(1.5\%)$ & $(1.1\%)$ & $(1.0\%)$ & $(0.9\%)$ & $(1.0\%)$ & $(0.3\%)$ \\
& VGG-11 & $4.5\%$ & $9.8\%$ & $18.9\%$ & $51.6\%$ & $14.9\%$ & $0.6\%$ \\
& & $(0.9\%)$ & $(2.8\%)$ & $(1.8\%)$ & $(1.8\%)$ & $(0.9\%)$ & $(0.2\%)$ \\
& ConvNext-T & $7.2\%$ & $2.6\%$ & $15.0\%$ & $62.4\%$ & $13.0\%$ & $0.2\%$ \\
& & $(2.2\%)$ & $(4.8\%)$ & $(3.6\%)$ & $(2.2\%)$ & $(1.6\%)$ & $(0.2\%)$ \\
\midrule
CelebA & DeiT-Ti & $2.0\%$ & $5.1\%$ & $8.7\%$ & $18.5\%$ & $58.9\%$ & $7.2\%$ \\
& & $(0.7\%)$ & $(2.1\%)$ & $(1.4\%)$ & $(0.6\%)$ & $(2.1\%)$ & $(0.6\%)$ \\
& ResNet-18 & $0.7\%$ & $-3.2\%$ & $9.2\%$ & $23.0\%$ & $64.5\%$ & $6.2\%$ \\
& & $(1.7\%)$ & $(1.4\%)$ & $(0.7\%)$ & $(1.8\%)$ & $(1.7\%)$ & $(1.4\%)$ \\
& VGG-11 & $4.1\%$ & $1.5\%$ & $7.1\%$ & $27.7\%$ & $60.0\%$ & $0.1\%$ \\
& & $(0.6\%)$ & $(0.6\%)$ & $(1.3\%)$ & $(1.9\%)$ & $(2.3\%)$ & $(0.4\%)$ \\
& ConvNext-T & $-0.7\%$ & $3.0\%$ & $3.0\%$ & $18.1\%$ & $59.4\%$ & $17.5\%$ \\
& & $(1.7\%)$ & $(1.2\%)$ & $(1.7\%)$ & $(2.2\%)$ & $(2.0\%)$ & $(1.0\%)$ \\
\bottomrule
CIFAR-10 & DeiT-Ti & $0.6\%$ & $1.1\%$ & $1.9\%$ & $5.2\%$ & $11.0\%$ & $80.6\%$ \\
(strong) & & $(0.5\%)$ & $(0.5\%)$ & $(0.4\%)$ & $(0.2\%)$ & $(0.5\%)$ & $(0.8\%)$ \\
& ResNet-18 & $0.2\%$ & $0.9\%$ & $1.8\%$ & $2.0\%$ & $19.3\%$ & $76.1\%$ \\
& & $(0.4\%)$ & $(0.3\%)$ & $(0.4\%)$ & $(0.4\%)$ & $(0.7\%)$ & $(0.5\%)$ \\
& VGG-11 & $-0.1\%$ & $0.2\%$ & $0.7\%$ & $2.2\%$ & $21.7\%$ & $75.7\%$ \\
& & $(0.3\%)$ & $(0.2\%)$ & $(0.2\%)$ & $(0.3\%)$ & $(0.3\%)$ & $(0.5\%)$ \\
& ConvNext-T & $-0.2\%$ & $1.6\%$ & $-0.4\%$ & $5.1\%$ & $13.5\%$ & $80.7\%$ \\
& & $(0.8\%)$ & $(0.6\%)$ & $(0.3\%)$ & $(0.6\%)$ & $(0.9\%)$ & $(0.9\%)$ \\
\midrule
Waterbirds & DeiT-Ti & $0.7\%$ & $0.5\%$ & $1.9\%$ & $5.6\%$ & $23.0\%$ & $68.6\%$ \\
& & $(0.4\%)$ & $(0.2\%)$ & $(0.5\%)$ & $(0.4\%)$ & $(0.9\%)$ & $(0.9\%)$ \\
& ResNet-18 & $0.6\%$ & $1.1\%$ & $0.5\%$ & $2.5\%$ & $28.6\%$ & $67.0\%$ \\
& & $(0.4\%)$ & $(0.2\%)$ & $(0.4\%)$ & $(0.6\%)$ & $(3.0\%)$ & $(3.2\%)$ \\
& VGG-11 & $0.3\%$ & $0.2\%$ & $-0.2\%$ & $2.8\%$ & $29.8\%$ & $67.3\%$ \\
& & $(0.3\%)$ & $(0.3\%)$ & $(0.5\%)$ & $(0.7\%)$ & $(1.0\%)$ & $(1.0\%)$ \\
& ConvNext-T & $0.6\%$ & $0.5\%$ & $0.5\%$ & $2.2\%$ & $16.6\%$ & $79.8\%$ \\
& & $(0.3\%)$ & $(0.6\%)$ & $(0.3\%)$ & $(0.2\%)$ & $(0.6\%)$ & $(0.6\%)$ \\
\midrule
CelebA & DeiT-Ti & $0.0\%$ & $0.4\%$ & $2.2\%$ & $2.1\%$ & $10.9\%$ & $84.7\%$ \\
& & $(0.1\%)$ & $(0.4\%)$ & $(0.5\%)$ & $(0.5\%)$ & $(0.7\%)$ & $(1.1\%)$ \\
& ResNet-18 & $-0.2\%$ & $0.1\%$ & $1.0\%$ & $3.3\%$ & $15.9\%$ & $80.2\%$ \\
& & $(0.4\%)$ & $(0.2\%)$ & $(0.4\%)$ & $(0.7\%)$ & $(1.0\%)$ & $(0.7\%)$ \\
& VGG-11 & $0.1\%$ & $0.4\%$ & $1.9\%$ & $4.6\%$ & $26.5\%$ & $66.9\%$ \\
& & $(0.3\%)$ & $(0.2\%)$ & $(0.2\%)$ & $(0.8\%)$ & $(2.4\%)$ & $(1.7\%)$ \\
& ConvNext-T & $-0.3\%$ & $-0.0\%$ & $-0.0\%$ & $2.0\%$ & $10.7\%$ & $87.9\%$ \\
& & $(0.3\%)$ & $(0.3\%)$ & $(0.2\%)$ & $(0.4\%)$ & $(0.6\%)$ & $(1.1\%)$ \\
\bottomrule
\\
    \end{tabular}
\end{table*}

\section{DISCUSSION}

We analyzed the localization of shortcuts in deep models. We decomposed
shortcut learning into two fundamental processes: spurious feature promotion
and core feature degradation. Using our counterfactual retraining method, we
examined them through two metrics: spurious feature encoding and core feature
forgetting.

Our findings demonstrate that neither encoding nor forgetting is localized in
any single layer within vision models. The interactions between layers play a
crucial role in shortcut formation. Earlier blocks typically facilitate
spurious feature encoding, while latter blocks are responsible for core feature
forgetting.

\paragraph{Practical Implications for Fine-tuning} Examing the axes of variation in our metrics, we found that dataset and model architecture play a crucial role in the
localization of encoding, while skew frequency and optimizer are
important for forgetting. Additionally, our
localization metrics are predictive of the success of some layer-wise interventions.
These results jointly suggest that shortcut mitigations based on layer-wise
manipulation of learning rates and frozen layers \citep[e.g.,][]{l23s} should
be dataset- and architecture-specific.

\paragraph{Future Work}
We hope our results will facilitate future research on developing more robust
models that can effectively resist spurious correlations.

Specifically, our observations suggest a trade-off between feature extractor
adaptability and robustness, which would be interesting to study in the future.
Feature extractors trained on clean data provide greater robustness against
shortcuts by compelling the final classifier layers to utilize core features.
However, models with clean feature extractors have higher error rate on fully
skewed datasets (see \cref{tab:c10m-s-r-s-err} in Appendix).

\clearpage

\section*{Acknowledgments}

This research was partially funded from the Ministry of Education and Science
of Bulgaria (support for INSAIT, part of the Bulgarian National Roadmap for
Research Infrastructure). This project was supported with computational
resources provided by Google Cloud Platform (GCP).

\bibliography{ta-bib}

\clearpage

\appendix

\thispagestyle{empty}

\onecolumn
\aistatstitle{Supplementary Material}


\startcontents
\printcontents{}{1}[2]{}

\section{ADDITIONAL RESULTS FOR FINE-TUNED MODELS}
\label{sec:add-results}

\paragraph{Localization-Guided Interventions} \cref{tab:full-regression}
present the full version of \cref{tab:regression}.

\begin{table}[htb]
    \centering
    \caption{Predictive power of localization metrics with regression
    coefficients (and their standard errors)}
    \label{tab:full-regression}
    \begin{tabular}{l c c c c c}
\toprule
& LR$\uparrow$ & LR$\downarrow$ & WD$\uparrow$ & WD$\downarrow$ & Freeze \\
\midrule
Enc & $-0.35$ & $0.20$ & $-0.02$ & $-0.02$ & $-1.36$ \\
& $(0.11)$ & $(0.07)$ & $(0.02)$ & $(0.02)$ & $(0.92)$ \\
Fgt & $0.39$ & $-0.45$ & $0.01$ & $0.01$ & $5.16$ \\
& $(0.11)$ & $(0.09)$ & $(0.02)$ & $(0.02)$ & $(1.31)$ \\
Enc $\times$ Fgt & $0.14$ & $0.13$ & $0.00$ & $0.01$ & $-6.25$ \\
& $(0.11)$ & $(0.10)$ & $(0.02)$ & $(0.02)$ & $(3.18)$ \\
Enc${}^2$ & $0.21$ & $-0.11$ & $0.01$ & $0.02$ & $2.25$ \\
& $(0.14)$ & $(0.08)$ & $(0.02)$ & $(0.02)$ & $(1.57)$ \\
Fgt${}^2$ & $-0.12$ & $-0.06$ & $-0.00$ & $-0.01$ & $-6.96$ \\
& $(0.14)$ & $(0.14)$ & $(0.02)$ & $(0.03)$ & $(2.20)$ \\
First & $-0.03$ & $0.03$ & $-0.00$ & $-0.00$ & $0.12$ \\
& $(0.02)$ & $(0.02)$ & $(0.01)$ & $(0.01)$ & $(0.17)$ \\
Last & $-0.22$ & $0.34$ & $-0.01$ & $-0.00$ & $0.12$ \\
& $(0.08)$ & $(0.08)$ & $(0.02)$ & $(0.02)$ & $(0.92)$ \\
Const & $-0.01$ & $-0.02$ & $0.00$ & $0.00$ & $-0.22$ \\
& $(0.02)$ & $(0.01)$ & $(0.01)$ & $(0.01)$ & $(0.13)$ \\
Double & $0.02$ & $-0.01$ & $0.00$ & $0.00$ \\
& $(0.02)$ & $(0.01)$ & $(0.01)$ & $(0.01)$ \\
\midrule
F-Stat & $5.04$ & $8.76$ & $0.66$ & $0.44$ & $4.07$ \\
R${}^2$ & $0.36$ & $0.54$ & $0.07$ & $0.05$ & $0.32$ \\
N & $99$ & $99$ & $99$ & $99$ & $54$ \\
\bottomrule
    \end{tabular}
\end{table}

\paragraph{Error Rates on Fully Skewed Dataset}
\cref{tab:c10m-s-r-s-err} presents error rates of the clean models intervened
with sets $A = i \mathbin{:} m$ for AdamW fine-tuned architectures on CIFAR-10
(strong) with common skew. As we can see, the error rates increase with the
number of initial clean blocks.

\begin{table}[htb]
    \centering
    \caption{Average test error rates (and standard errors) on the fully
    skewed CIFAR-10 (strong) dataset of clean models intervened with sets $A =
    i \mathbin{:} m$ for AdamW fine-tuned architectures on common skew}
    \label{tab:c10m-s-r-s-err}
    \begin{tabular}{c c c c c c c c}
\toprule
Model & Skewed & $1\mathbin{:}6$ & $2\mathbin{:}6$ & $3\mathbin{:}6$ & $4\mathbin{:}6$ & $5\mathbin{:}6$ & Clean\\
\midrule
DeiT-Ti & $0.3\%$ & $0.3\%$ & $0.5\%$ & $1.3\%$ & $2.3\%$ & $3.4\%$ & $3.4\%$ \\
& $(0.1\%)$ & $(0.1\%)$ & $(0.1\%)$ & $(0.1\%)$ & $(0.1\%)$ & $(0.1\%)$ & $(0.1\%)$ \\
ResNet-18 & $0.3\%$ & $0.3\%$ & $0.3\%$ & $0.4\%$ & $1.3\%$ & $4.2\%$ & $4.2\%$ \\
& $(0.1\%)$ & $(0.1\%)$ & $(0.1\%)$ & $(0.1\%)$ & $(0.1\%)$ & $(0.1\%)$ & $(0.1\%)$ \\
VGG-11 & $0.3\%$ & $0.3\%$ & $0.3\%$ & $0.3\%$ & $2.4\%$ & $7.2\%$ & $7.2\%$ \\
& $(0.1\%)$ & $(0.1\%)$ & $(0.1\%)$ & $(0.1\%)$ & $(0.1\%)$ & $(0.1\%)$ & $(0.1\%)$ \\
\bottomrule
    \end{tabular}
\end{table}

\paragraph{Additional Results on Localization in a Single Block}
\cref{tab:ca-wb-ind} follows \cref{tab:cifar-ind} for the models fine-tuned
with AdamW on CelebA and Waterbirds with common skew. Similarly to the
CIFAR-10 results, there does not exist an individual layer fully responsible
for shortcut learning. As previously, the sum of individual contributions to
forgetting does not reach $100\%$, which suggest that forgetting can not be
explained by the sum of individual contributions. In contrast to the previous
results, the sum of individual contributions to encoding does not reach $100\%$
for CelebA and approximately equals $100\%$ for Waterbirds. These results
again suggest that the sum of individual contributions can not reliably explain
spurious feature encoding. While this approach gives plausible results for the
Waterbirds dataset, it fails for the CIFAR-10 and CelebA datasets.

\begin{table}[htb]
    \centering
    \caption{Average relative contributions (and standard errors) of single
    blocks to encoding (first and third sub-part) and forgetting (second and
    fourth sub-parts) for models fine-tuned with AdamW on CelebA (top part) and
    Waterbirds (bottom part) with common skew}
    \label{tab:ca-wb-ind}
    \begin{tabular}{l c c c c c c}
\toprule
Model & Bl. $0$ & Bl. $1$ & Bl. $2$ & Bl. $3$ & Bl. $4$ & Bl. $5$ \\
\midrule
DeiT-Ti & $2.0\%$ & $3.7\%$ & $6.6\%$ & $-0.1\%$ & $2.1\%$ & $0.7\%$ \\
& $(0.7\%)$ & $(1.9\%)$ & $(1.9\%)$ & $(1.5\%)$ & $(1.3\%)$ & $(0.6\%)$ \\
ResNet-18 & $0.7\%$ & $-1.4\%$ & $0.7\%$ & $-11.7\%$ & $12.1\%$ & $-0.1\%$ \\
& $(1.7\%)$ & $(5.9\%)$ & $(3.1\%)$ & $(3.2\%)$ & $(5.6\%)$ & $(1.0\%)$ \\
VGG-11 & $4.1\%$ & $3.2\%$ & $4.9\%$ & $5.0\%$ & $-73.3\%$ & $-4.3\%$ \\
& $(0.6\%)$ & $(1.0\%)$ & $(2.4\%)$ & $(6.8\%)$ & $(10.6\%)$ & $(0.7\%)$ \\
ConvNeXt-Ti & $0.2\%$ & $2.6\%$ & $2.2\%$ & $9.8\%$ & $-0.4\%$ & $-2.6\%$ \\
& $(1.8\%)$ & $(1.7\%)$ & $(2.3\%)$ & $(3.6\%)$ & $(2.6\%)$ & $(1.1\%)$ \\
\midrule
DeiT-Ti & $0.0\%$ & $0.2\%$ & $0.9\%$ & $0.5\%$ & $0.2\%$ & $0.0\%$ \\
& $(0.1\%)$ & $(0.2\%)$ & $(0.2\%)$ & $(0.2\%)$ & $(0.3\%)$ & $(0.2\%)$ \\
ResNet-18 & $-0.2\%$ & $0.8\%$ & $0.4\%$ & $2.3\%$ & $23.8\%$ & $-0.1\%$ \\
& $(0.4\%)$ & $(0.5\%)$ & $(0.3\%)$ & $(0.4\%)$ & $(13.6\%)$ & $(0.2\%)$ \\
VGG-11 & $0.1\%$ & $0.3\%$ & $1.7\%$ & $7.9\%$ & $0.6\%$ & $-0.1\%$ \\
& $(0.3\%)$ & $(0.3\%)$ & $(0.7\%)$ & $(2.2\%)$ & $(0.7\%)$ & $(0.3\%)$ \\
ConvNeXt-Ti & $-0.0\%$ & $-0.4\%$ & $-0.2\%$ & $1.4\%$ & $0.1\%$ & $-0.0\%$ \\
& $(0.3\%)$ & $(0.2\%)$ & $(0.4\%)$ & $(0.7\%)$ & $(0.5\%)$ & $(0.2\%)$ \\
\bottomrule
DeiT-Ti & $1.2\%$ & $16.7\%$ & $32.4\%$ & $31.2\%$ & $14.8\%$ & $3.2\%$ \\
& $(2.4\%)$ & $(2.1\%)$ & $(4.8\%)$ & $(2.4\%)$ & $(2.6\%)$ & $(1.5\%)$ \\
ResNet-18 & $6.7\%$ & $12.4\%$ & $24.8\%$ & $40.7\%$ & $17.0\%$ & $1.4\%$ \\
& $(1.5\%)$ & $(1.8\%)$ & $(1.9\%)$ & $(3.5\%)$ & $(7.0\%)$ & $(0.5\%)$ \\
VGG-11 & $4.5\%$ & $10.2\%$ & $31.3\%$ & $64.3\%$ & $18.9\%$ & $2.0\%$ \\
& $(0.9\%)$ & $(2.9\%)$ & $(0.6\%)$ & $(2.3\%)$ & $(4.7\%)$ & $(0.5\%)$ \\
ConvNeXt-Ti & $-3.1\%$ & $2.4\%$ & $12.7\%$ & $77.7\%$ & $51.3\%$ & $1.7\%$ \\
& $(5.3\%)$ & $(2.0\%)$ & $(4.7\%)$ & $(3.1\%)$ & $(1.3\%)$ & $(4.7\%)$ \\
\midrule
DeiT-Ti & $0.7\%$ & $0.7\%$ & $1.1\%$ & $1.8\%$ & $0.9\%$ & $0.1\%$ \\
& $(0.4\%)$ & $(0.4\%)$ & $(0.3\%)$ & $(0.4\%)$ & $(0.4\%)$ & $(0.4\%)$ \\
ResNet-18 & $0.6\%$ & $0.5\%$ & $0.7\%$ & $1.1\%$ & $20.7\%$ & $0.1\%$ \\
& $(0.4\%)$ & $(0.4\%)$ & $(0.2\%)$ & $(0.5\%)$ & $(3.7\%)$ & $(0.3\%)$ \\
VGG-11 & $0.3\%$ & $0.2\%$ & $1.1\%$ & $7.0\%$ & $1.4\%$ & $0.2\%$ \\
& $(0.3\%)$ & $(0.2\%)$ & $(0.6\%)$ & $(1.1\%)$ & $(0.2\%)$ & $(0.1\%)$ \\
ConvNeXt-Ti & $1.5\%$ & $1.2\%$ & $1.1\%$ & $1.8\%$ & $1.5\%$ & $0.6\%$ \\
& $(1.0\%)$ & $(0.8\%)$ & $(0.6\%)$ & $(0.4\%)$ & $(0.7\%)$ & $(0.3\%)$ \\
\bottomrule
    \end{tabular}
\end{table}

Interestingly, Block 4 of the VGG-11 model exhibits a strong negative
contribution to encoding. To understand this behavior, we examined the
individual contributions to encoding for the same models trained on rare skew
in \cref{tab:ca-r-ind} and the error rates of these models in
\cref{tab:vgg-celeba-s-err}. Similarly to the common skew, Block 4 of the
VGG-11 model exhibits a strong negative contribution to encoding. Also, all
intervened models achieve a small error rate on the fully skewed dataset. A
plausible explanation of this behavior is the following. Block 4 in the clean
model is not well suited for spurious feature encoding. Due to regularization
(i.e., implicit biases of optimizer and weight decay), the complement to this
block can not overcome this restriction and starts to mix
core features with spurious features, which corrupts the core features and
leads to a significant accuracy drop on the clean dataset. This example again
suggests that shortcut learning crucially depends on the interactions of
different layers within the network.

\begin{table}[htb]
    \centering
    \caption{Average relative contributions (and standard errors) of single
    blocks to encoding for models fine-tuned with AdamW on CelebA with rare
    skew}
    \label{tab:ca-r-ind}
    \begin{tabular}{l c c c c c c}
\toprule
Model & Bl. $0$ & Bl. $1$ & Bl. $2$ & Bl. $3$ & Bl. $4$ & Bl. $5$ \\
\midrule
VGG-11 & $4.1\%$ & $2.2\%$ & $-3.6\%$ & $-6.1\%$ & $-56.7\%$ & $-1.9\%$ \\
& $(0.9\%)$ & $(1.6\%)$ & $(2.1\%)$ & $(7.6\%)$ & $(12.7\%)$ & $(0.7\%)$ \\
\bottomrule
    \end{tabular}
\end{table}

\begin{table}[htb]
    \centering
    \caption{Average test error rates (and standard errors) on the clean (first
    and third row) and fully skewed (second and fourth row) CelebA dataset
    of clean models intervened with sets $A = [m] \setminus \{i\}$ for VGG-11
    architecture fine-tuned with AdamW on common (top) and rare (bottom) skews}
    \label{tab:vgg-celeba-s-err}
    \begin{tabular}{c c c c c c c c}
\toprule
Skewed & $\{-0\}$ & $\{-1\}$ & $\{-2\}$ & $\{-3\}$ & $\{-4\}$ & $\{-5\}$ & Clean\\
\midrule
$21.3\%$ & $20.7\%$ & $20.8\%$ & $20.6\%$ & $20.5\%$ & $32.5\%$ & $22.0\%$ & $6.0\%$ \\
$(0.7\%)$ & $(0.6\%)$ & $(0.6\%)$ & $(0.9\%)$ & $(0.9\%)$ & $(1.9\%)$ & $(0.6\%)$ & $(0.1\%)$ \\
$0.31\%$ & $0.31\%$ & $0.31\%$ & $0.35\%$ & $0.50\%$ & $0.49\%$ & $0.31\%$ & $5.09\%$ \\
$(0.01\%)$ & $(0.02\%)$ & $(0.02\%)$ & $(0.02\%)$ & $(0.06\%)$ & $(0.05\%)$ & $(0.01\%)$ & $(0.04\%)$ \\
\midrule
$11.0\%$ & $10.8\%$ & $10.9\%$ & $11.2\%$ & $11.2\%$ & $13.7\%$ & $11.1\%$ & $6.0\%$ \\
$(0.3\%)$ & $(0.3\%)$ & $(0.4\%)$ & $(0.4\%)$ & $(0.2\%)$ & $(0.7\%)$ & $(0.4\%)$ & $(0.1\%)$ \\
$0.64\%$ & $0.67\%$ & $0.66\%$ & $0.67\%$ & $0.75\%$ & $0.56\%$ & $0.63\%$ & $5.09\%$ \\
$(0.04\%)$ & $(0.04\%)$ & $(0.05\%)$ & $(0.06\%)$ & $(0.05\%)$ & $(0.01\%)$ & $(0.04\%)$ & $(0.04\%)$ \\
\bottomrule
    \end{tabular}
\end{table}

\clearpage

\section{TRAINING FROM SCRATCH}
\label{sec:scratch}

This section presents the general trends for models trained from scratch
observed in our experiments. We didn't train ConvNeXt-T model from scratch due
to computational constraints. Also, note that we did not manage to train the
DeiT-Ti architecture on the Waterbirds dataset from scratch. Thus, experiments
for this pair are omitted. As previously, we first report the error rates
achieved by models on different datasets in \cref{tab:err-rates-s}. As we can
see, shortcut learning is exacerbated for models trained from scratch. Also, we
see that DeiT models have significantly higher error rates on CIFAR-10 because,
generally, ViT architectures are more ``data hungry'' \citep{z23u}. Also, we
can see that AdamW achieves better error rates compared to SGD for ViT
architectures, while the opposite is true for CNN architectures.

\begin{table}[htb]
    \centering
    \caption{Average clean test error rates (and their standard errors in
    parenthesis) of clean and skewed models trained from scratch with AdamW
    (top part) and SGD (bottom part) on rare (first and third sub-parts) and
    common (second and fourth sub-parts) skews}
    \label{tab:err-rates-s}
    \begin{tabular}{l c c c c c c c c}
\toprule
& \multicolumn{2}{c}{CIFAR-10 (weak)} & \multicolumn{2}{c}{CIFAR-10 (strong)} & \multicolumn{2}{c}{Waterbirds} & \multicolumn{2}{c}{CelebA} \\
\cmidrule(lr){2-3} \cmidrule(lr){4-5} \cmidrule(lr){6-7} \cmidrule(lr){8-9}
Model & Clean & Skewed & Clean & Skewed & Clean & Skewed & Clean & Skewed \\
\midrule
DeiT-Ti & $18.7\%$ & $20.8\%$ & $18.9\%$ & $41.8\%$ & $-$ & $-$ & $7.1\%$ & $13.3\%$ \\
& $(0.2\%)$ & $(0.3\%)$ & $(0.2\%)$ & $(0.4\%)$ & $-$ & $-$ & $(0.1\%)$ & $(0.1\%)$ \\
ResNet-18 & $6.7\%$ & $19.8\%$ & $7.0\%$ & $25.6\%$ & $7.1\%$ & $26.9\%$ & $6.3\%$ & $11.3\%$ \\
& $(0.2\%)$ & $(0.2\%)$ & $(0.1\%)$ & $(0.2\%)$ & $(0.2\%)$ & $(0.5\%)$ & $(0.1\%)$ & $(0.4\%)$ \\
VGG-11 & $6.7\%$ & $24.2\%$ & $7.0\%$ & $28.7\%$ & $4.9\%$ & $25.0\%$ & $6.3\%$ & $11.2\%$ \\
& $(0.1\%)$ & $(0.3\%)$ & $(0.1\%)$ & $(0.3\%)$ & $(0.2\%)$ & $(0.6\%)$ & $(0.1\%)$ & $(0.4\%)$ \\
\midrule
DeiT-Ti & $18.4\%$ & $25.2\%$ & $18.7\%$ & $63.3\%$ & $-$ & $-$ & $7.1\%$ & $22.3\%$ \\
& $(0.2\%)$ & $(0.6\%)$ & $(0.1\%)$ & $(0.3\%)$ & $-$ & $-$ & $(0.1\%)$ & $(0.3\%)$ \\
ResNet-18 & $6.7\%$ & $35.1\%$ & $7.0\%$ & $48.4\%$ & $7.1\%$ & $36.7\%$ & $6.3\%$ & $21.4\%$ \\
& $(0.2\%)$ & $(0.4\%)$ & $(0.1\%)$ & $(0.3\%)$ & $(0.2\%)$ & $(0.4\%)$ & $(0.1\%)$ & $(0.6\%)$ \\
VGG-11 & $6.7\%$ & $46.1\%$ & $7.0\%$ & $57.5\%$ & $4.9\%$ & $35.2\%$ & $6.3\%$ & $20.5\%$ \\
& $(0.1\%)$ & $(0.5\%)$ & $(0.1\%)$ & $(0.4\%)$ & $(0.2\%)$ & $(0.5\%)$ & $(0.1\%)$ & $(0.3\%)$ \\
\midrule
\midrule
DeiT-Ti & $27.5\%$ & $28.2\%$ & $27.6\%$ & $44.1\%$ & $-$ & $-$ & $8.7\%$ & $12.7\%$ \\
& $(0.3\%)$ & $(0.2\%)$ & $(0.2\%)$ & $(0.3\%)$ & $-$ & $-$ & $(0.2\%)$ & $(0.2\%)$ \\
ResNet-18 & $5.6\%$ & $18.3\%$ & $5.9\%$ & $24.5\%$ & $6.5\%$ & $27.3\%$ & $6.3\%$ & $11.4\%$ \\
& $(0.1\%)$ & $(0.1\%)$ & $(0.1\%)$ & $(0.2\%)$ & $(0.2\%)$ & $(0.8\%)$ & $(0.1\%)$ & $(0.2\%)$ \\
VGG-11 & $6.4\%$ & $23.1\%$ & $6.8\%$ & $28.6\%$ & $4.9\%$ & $25.4\%$ & $6.3\%$ & $11.2\%$ \\
& $(0.1\%)$ & $(0.1\%)$ & $(0.2\%)$ & $(0.2\%)$ & $(0.1\%)$ & $(0.5\%)$ & $(0.1\%)$ & $(0.2\%)$ \\
\midrule
DeiT-Ti & $27.8\%$ & $28.7\%$ & $27.4\%$ & $59.1\%$ & $-$ & $-$ & $8.8\%$ & $14.9\%$ \\
& $(0.3\%)$ & $(0.3\%)$ & $(0.3\%)$ & $(0.3\%)$ & $-$ & $-$ & $(0.2\%)$ & $(0.3\%)$ \\
ResNet-18 & $5.6\%$ & $32.6\%$ & $5.9\%$ & $47.9\%$ & $6.5\%$ & $36.5\%$ & $6.3\%$ & $22.0\%$ \\
& $(0.1\%)$ & $(0.4\%)$ & $(0.1\%)$ & $(0.3\%)$ & $(0.2\%)$ & $(0.4\%)$ & $(0.1\%)$ & $(0.4\%)$ \\
VGG-11 & $6.4\%$ & $43.8\%$ & $6.8\%$ & $57.0\%$ & $4.9\%$ & $35.2\%$ & $6.3\%$ & $19.9\%$ \\
& $(0.1\%)$ & $(0.5\%)$ & $(0.2\%)$ & $(0.5\%)$ & $(0.1\%)$ & $(0.6\%)$ & $(0.1\%)$ & $(0.2\%)$ \\
\bottomrule
    \end{tabular}
\end{table}

\paragraph{Localization in a Single Block}
\cref{tab:cifar-ind-s} replicates \cref{tab:cifar-ind} for the models trained
from scratch. In this experiment, we observed divergence in some intervened
VGG-11 models: contributions of the corresponding models are not included in
the mean calculation.\footnote{Similarly to the fine-tuning case, we did not
observe divergence in the initial blocks setting. In the single block setting,
we observed divergence in 227 out of 2640 models.} Generally, we observe trends
similar to those previously observed with the fine-tuned models.

\begin{table}[htb]
    \centering
    \caption{Average relative contributions (and standard errors) of single
    blocks to encoding (top) and forgetting (bottom) for models trained from
    scratch with AdamW on CIFAR-10 (strong) with common skew}
    \label{tab:cifar-ind-s}
    \begin{tabular}{l c c c c c c}
\toprule
Model & Bl. $0$ & Bl. $1$ & Bl. $2$ & Bl. $3$ & Bl. $4$ & Bl. $5$ \\
\midrule
DeiT-Ti & $24.1\%$ & $9.0\%$ & $13.7\%$ & $14.2\%$ & $11.6\%$ & $-1.1\%$ \\
& $(1.0\%)$ & $(0.6\%)$ & $(0.9\%)$ & $(1.1\%)$ & $(1.1\%)$ & $(0.8\%)$ \\
ResNet-18 & $0.3\%$ & $-17.8\%$ & $-23.5\%$ & $-0.6\%$ & $19.3\%$ & $0.6\%$ \\
& $(1.1\%)$ & $(6.3\%)$ & $(9.1\%)$ & $(4.7\%)$ & $(2.1\%)$ & $(1.1\%)$ \\
VGG-11 & $3.4\%$ & $-$ & $-24.2\%$ & $25.9\%$ & $16.9\%$ & $0.0\%$ \\
& $(0.7\%)$ & $-$ & $-$ & $(2.8\%)$ & $(3.4\%)$ & $(0.6\%)$ \\
\midrule
DeiT-Ti & $-0.4\%$ & $-1.4\%$ & $0.7\%$ & $0.3\%$ & $0.5\%$ & $-0.2\%$ \\
& $(0.2\%)$ & $(0.5\%)$ & $(0.5\%)$ & $(0.4\%)$ & $(0.4\%)$ & $(0.4\%)$ \\
ResNet-18 & $0.8\%$ & $2.7\%$ & $21.8\%$ & $18.7\%$ & $23.3\%$ & $-0.1\%$ \\
& $(0.3\%)$ & $(0.4\%)$ & $(6.2\%)$ & $(2.2\%)$ & $(2.1\%)$ & $(0.3\%)$ \\
VGG-11 & $0.7\%$ & $61.2\%$ & $63.7\%$ & $57.1\%$ & $-$ & $1.4\%$ \\
& $(0.4\%)$ & $(21.7\%)$ & $(28.2\%)$ & $(12.6\%)$ & $-$ & $(0.2\%)$ \\
\bottomrule
    \end{tabular}
\end{table}

\paragraph{Localization in the Initial Blocks}
\cref{tab:init-adamw-s} follows \cref{tab:init-adamw} for the models trained
from scratch. Compared to fine-tuned models, the first layer of transformer
models starts to play an important role in spurious feature encoding. At the
same time, ViT and CNN architectures demonstrate the opposite trends in
spurious feature encoding. Specifically, for CIFAR-10 and CelebA, ViT
architectures tend to encode the spurious feature earlier when trained from
scratch compared to fine-tuning, while CNN architectures tend to encode the
spurious feature later. However, models trained from scratch seem to encode the
background skew in earlier layers compared to the fine-tuned models.

Additionally, all architectures tend to forget the core feature in the latter
layers when trained from scratch (however, this trend is less pronounced for
ViT architectures). Importantly, the trends about the effects of models on
spurious feature encoding also seem to hold for the models trained from
scratch. As for the effect of datasets, the watermark skew is still encoded
earlier than the sampling skew. However, the background skew is now encoded
earlier than the watermark skew.

\cref{tab:forget-decomp-s} follows \cref{tab:forget-decomp} for the models
trained from scratch. Generally, forgetting seems to be even more concentrated
in the last and penultimate blocks for models trained from scratch. Also,
similarly to the fine-tuned models, common skews seem to induce forgetting in
the latter layers. However, it is hard to see more specific trends.

\paragraph{Main Explanatory Factors}
\cref{tab:var-decomp-s} follows \cref{tab:var-decomp} for the models trained
from scratch. While the results for the encoding are similar between fine-tuned
models and models trained from scratch, the main explanatory factors for the
forgetting shift to dataset and model architecture suggesting that forgetting
in the models trained from scratch follows different mechanisms compared to
fine-tuned models. Dataset and architecture together explain $79.4\%$ and
$51.8\%$ of variance in the localization of encoding and forgetting,
respectively.

\begin{table}[htb]
    \centering
    \caption{Variance explained in relative encoding (left) and forgetting
    (right) rate by different factors}
    \label{tab:var-decomp-s}
    \begin{tabular}{c c c c c c c c}
\toprule
\multicolumn{4}{c}{Encoding} & \multicolumn{4}{c}{Forgetting} \\
\cmidrule(lr){1-4} \cmidrule(lr){5-8}
Dataset & Skew freq. & Model & Optimizer & Dataset & Skew freq. & Model & Optimizer\\
\midrule
$40.7\%$ & $0.4\%$ & $23.1\%$ & $0.6\%$ & $18.8\%$ & $4.6\%$ & $13.7\%$ & $2.1\%$ \\
\bottomrule
    \end{tabular}
\end{table}

\begin{table}[htb]
    \centering
    \caption{Average increase rate in relative contributions (and standard
    errors) of the initial blocks to encoding (top) and forgetting (bottom) for
    models fine-tuned with AdamW on common skews}
    \label{tab:init-adamw-s}
    \begin{tabular}{l l c c c c c c}
\toprule
Dataset & Model & Bl. $0$ & Bl. $1$ & Bl. $2$ & Bl. $3$ & Bl. $4$ & Bl. $5$ \\
\midrule
CIFAR-10 & DeiT-Ti & $75.3\%$ & $2.9\%$ & $14.2\%$ & $6.1\%$ & $1.5\%$ & $0.2\%$ \\
(weak) & & $(3.8\%)$ & $(0.9\%)$ & $(2.4\%)$ & $(2.5\%)$ & $(1.6\%)$ & $(0.5\%)$ \\
& ResNet-18 & $2.7\%$ & $5.4\%$ & $14.5\%$ & $49.0\%$ & $28.6\%$ & $0.1\%$ \\
& & $(1.3\%)$ & $(0.9\%)$ & $(1.0\%)$ & $(2.0\%)$ & $(0.7\%)$ & $(0.1\%)$ \\
& VGG-11 & $1.4\%$ & $4.1\%$ & $12.0\%$ & $36.8\%$ & $45.9\%$ & $0.0\%$ \\
& & $(1.2\%)$ & $(1.0\%)$ & $(1.4\%)$ & $(0.5\%)$ & $(1.0\%)$ & $(0.1\%)$ \\
\midrule
CIFAR-10 & DeiT-Ti & $23.7\%$ & $18.9\%$ & $23.9\%$ & $20.6\%$ & $13.1\%$ & $0.1\%$ \\
(strong) & & $(1.7\%)$ & $(1.2\%)$ & $(1.5\%)$ & $(1.2\%)$ & $(0.7\%)$ & $(0.1\%)$ \\
& ResNet-18 & $0.3\%$ & $4.0\%$ & $14.1\%$ & $48.9\%$ & $32.9\%$ & $0.1\%$ \\
& & $(1.1\%)$ & $(1.0\%)$ & $(1.5\%)$ & $(1.2\%)$ & $(1.1\%)$ & $(0.1\%)$ \\
& VGG-11 & $3.4\%$ & $-0.5\%$ & $13.6\%$ & $31.8\%$ & $51.9\%$ & $0.1\%$ \\
& & $(0.7\%)$ & $(1.0\%)$ & $(1.4\%)$ & $(1.3\%)$ & $(0.5\%)$ & $(0.1\%)$ \\
\midrule
Waterbirds & ResNet-18 & $17.2\%$ & $7.7\%$ & $21.4\%$ & $26.5\%$ & $26.3\%$ & $1.3\%$ \\
& & $(1.8\%)$ & $(1.8\%)$ & $(1.0\%)$ & $(1.4\%)$ & $(1.1\%)$ & $(0.6\%)$ \\
& VGG-11 & $10.5\%$ & $12.1\%$ & $33.0\%$ & $28.2\%$ & $15.6\%$ & $0.9\%$ \\
& & $(0.8\%)$ & $(1.4\%)$ & $(1.9\%)$ & $(1.1\%)$ & $(0.7\%)$ & $(0.6\%)$ \\
\midrule
CelebA & DeiT-Ti & $20.3\%$ & $14.0\%$ & $11.1\%$ & $13.1\%$ & $39.0\%$ & $2.8\%$ \\
& & $(2.6\%)$ & $(1.8\%)$ & $(2.7\%)$ & $(2.3\%)$ & $(1.9\%)$ & $(0.2\%)$ \\
& ResNet-18 & $2.1\%$ & $2.4\%$ & $6.6\%$ & $26.4\%$ & $58.5\%$ & $4.4\%$ \\
& & $(3.7\%)$ & $(1.7\%)$ & $(2.7\%)$ & $(4.5\%)$ & $(5.5\%)$ & $(1.0\%)$ \\
& VGG-11 & $-3.3\%$ & $6.5\%$ & $5.6\%$ & $18.3\%$ & $67.0\%$ & $6.2\%$ \\
& & $(3.7\%)$ & $(2.5\%)$ & $(2.7\%)$ & $(0.9\%)$ & $(1.2\%)$ & $(0.5\%)$ \\
\midrule
\midrule
CIFAR-10 & DeiT-Ti & $1.7\%$ & $-11.0\%$ & $-2.1\%$ & $4.6\%$ & $29.3\%$ & $77.8\%$ \\
(weak) & & $(3.4\%)$ & $(2.8\%)$ & $(3.2\%)$ & $(1.5\%)$ & $(4.5\%)$ & $(4.6\%)$ \\
& ResNet-18 & $0.6\%$ & $0.3\%$ & $1.5\%$ & $5.1\%$ & $15.1\%$ & $77.8\%$ \\
& & $(0.6\%)$ & $(0.5\%)$ & $(0.6\%)$ & $(0.3\%)$ & $(0.4\%)$ & $(0.2\%)$ \\
& VGG-11 & $0.8\%$ & $0.5\%$ & $3.1\%$ & $3.4\%$ & $12.7\%$ & $79.8\%$ \\
& & $(0.3\%)$ & $(0.3\%)$ & $(0.1\%)$ & $(0.3\%)$ & $(0.3\%)$ & $(0.3\%)$ \\
\midrule
CIFAR-10 & DeiT-Ti & $-0.6\%$ & $0.0\%$ & $3.7\%$ & $7.3\%$ & $15.1\%$ & $74.9\%$ \\
(strong) & & $(0.3\%)$ & $(0.4\%)$ & $(0.6\%)$ & $(1.3\%)$ & $(0.8\%)$ & $(1.0\%)$ \\
& ResNet-18 & $0.8\%$ & $0.6\%$ & $2.3\%$ & $5.1\%$ & $17.5\%$ & $73.9\%$ \\
& & $(0.3\%)$ & $(0.4\%)$ & $(0.3\%)$ & $(0.4\%)$ & $(0.6\%)$ & $(0.6\%)$ \\
& VGG-11 & $0.7\%$ & $1.4\%$ & $2.6\%$ & $4.3\%$ & $16.2\%$ & $75.1\%$ \\
& & $(0.4\%)$ & $(0.4\%)$ & $(0.2\%)$ & $(0.2\%)$ & $(0.3\%)$ & $(0.3\%)$ \\
\midrule
Waterbirds & ResNet-18 & $3.3\%$ & $1.1\%$ & $5.7\%$ & $13.2\%$ & $11.3\%$ & $65.7\%$ \\
& & $(0.6\%)$ & $(1.2\%)$ & $(1.2\%)$ & $(1.3\%)$ & $(1.6\%)$ & $(1.3\%)$ \\
& VGG-11 & $1.9\%$ & $2.4\%$ & $5.6\%$ & $11.9\%$ & $16.1\%$ & $62.3\%$ \\
& & $(0.6\%)$ & $(0.4\%)$ & $(0.4\%)$ & $(1.3\%)$ & $(1.0\%)$ & $(1.1\%)$ \\
\midrule
CelebA & DeiT-Ti & $0.5\%$ & $2.0\%$ & $-0.3\%$ & $1.9\%$ & $13.0\%$ & $83.2\%$ \\
& & $(0.5\%)$ & $(0.3\%)$ & $(0.2\%)$ & $(0.3\%)$ & $(1.1\%)$ & $(1.0\%)$ \\
& ResNet-18 & $0.6\%$ & $-0.7\%$ & $1.1\%$ & $3.2\%$ & $6.9\%$ & $89.3\%$ \\
& & $(0.6\%)$ & $(0.5\%)$ & $(0.4\%)$ & $(0.5\%)$ & $(0.9\%)$ & $(0.7\%)$ \\
& VGG-11 & $0.0\%$ & $0.4\%$ & $1.0\%$ & $3.0\%$ & $9.8\%$ & $86.1\%$ \\
& & $(0.4\%)$ & $(0.3\%)$ & $(0.5\%)$ & $(0.4\%)$ & $(0.9\%)$ & $(1.4\%)$ \\
\bottomrule
    \end{tabular}
\end{table}

\begin{table}[htb]
    \centering
    \caption{Average increase rate in relative forgetting (and standard errors)
    of the initial blocks of DeiT-Ti (top) and ResNet-18 (bottom) models
    trained from scratch on CIFAR-10 (strong)}
    \label{tab:forget-decomp-s}
    \begin{tabular}{l l c c c c c c}
\toprule
Frequency & Optimizer & Bl. $0$ & Bl. $1$ & Bl. $2$ & Bl. $3$ & Bl. $4$ & Bl. $5$ \\
\midrule
Rare & AdamW & $-1.4\%$ & $-1.5\%$ & $4.2\%$ & $9.4\%$ & $18.3\%$ & $71.3\%$ \\
& & $(1.4\%)$ & $(1.1\%)$ & $(0.9\%)$ & $(1.6\%)$ & $(1.1\%)$ & $(1.6\%)$ \\
& SGD & $-8.8\%$ & $-1.5\%$ & $1.0\%$ & $5.4\%$ & $26.7\%$ & $77.4\%$ \\
& & $(1.5\%)$ & $(1.6\%)$ & $(1.0\%)$ & $(0.7\%)$ & $(0.5\%)$ & $(1.4\%)$ \\
\midrule
Common & AdamW & $-0.6\%$ & $0.0\%$ & $3.7\%$ & $7.3\%$ & $15.1\%$ & $74.9\%$ \\
& & $(0.3\%)$ & $(0.4\%)$ & $(0.6\%)$ & $(1.3\%)$ & $(0.8\%)$ & $(1.0\%)$ \\
& SGD & $-4.7\%$ & $-1.7\%$ & $0.6\%$ & $3.6\%$ & $19.0\%$ & $83.6\%$ \\
& & $(1.1\%)$ & $(0.8\%)$ & $(0.3\%)$ & $(0.5\%)$ & $(0.8\%)$ & $(1.2\%)$ \\
\midrule
\midrule
Rare & AdamW & $1.1\%$ & $0.8\%$ & $3.7\%$ & $7.1\%$ & $20.7\%$ & $66.8\%$ \\
& & $(0.8\%)$ & $(0.8\%)$ & $(0.4\%)$ & $(0.4\%)$ & $(0.6\%)$ & $(0.5\%)$ \\
& SGD & $1.9\%$ & $1.4\%$ & $4.4\%$ & $6.6\%$ & $26.9\%$ & $59.2\%$ \\
& & $(0.8\%)$ & $(0.7\%)$ & $(0.6\%)$ & $(0.5\%)$ & $(0.5\%)$ & $(0.5\%)$ \\
\midrule
Common & AdamW & $0.8\%$ & $0.6\%$ & $2.3\%$ & $5.1\%$ & $17.5\%$ & $73.9\%$ \\
& & $(0.3\%)$ & $(0.4\%)$ & $(0.3\%)$ & $(0.4\%)$ & $(0.6\%)$ & $(0.6\%)$ \\
& SGD & $0.8\%$ & $1.2\%$ & $2.5\%$ & $4.5\%$ & $21.4\%$ & $70.0\%$ \\
& & $(0.2\%)$ & $(0.2\%)$ & $(0.2\%)$ & $(0.2\%)$ & $(0.5\%)$ & $(0.4\%)$ \\
\bottomrule
    \end{tabular}
\end{table}

\clearpage

\section{DETAILS OF TRAINING}
\label{sec:train-details}

We use the standard AdamW and SGD (with Nesterov momentum) optimizers from
PyTorch and cosine learning scheduler with linear warm-up. \cref{tab:epochs}
reports the number of training epochs. The hyper-parameters of optimizers are
listed in \cref{tab:fine-tune-params} (for fine-tuning) and
\cref{tab:scratch-params} (for training from scratch). For training, we use
standard augmentations: random resized crop and random horizontal flip. For
CelebA and Waterbirds, we use the same augmentation parameters as \citet{s20d}.
For CIFAR, we use scale $(0.8, 1.0)$ and ratio $(\nicefrac{3}{4},
\nicefrac{4}{3})$. We resize all images to size $224 \times 224$ for both
training and evaluation. For ResNet-18, VGG-11, and ConvNeXt-T fine-tuning, we
used the default weights from the TorchVision \citep{t16} library. For DeiT-Ti
fine-tuning, we used the default weights from the timm \citep{r19t} library.

We use the standard \texttt{train} CIFAR-10 and CUB-200-2011 \citep{w11c}
splits for the training on CIFAR-10 and Waterbirds.  We use the union of
\texttt{train} and \texttt{validation} splits for the training on CelebA. We
use \texttt{test} splits of the considered datasets for the evaluation. To make
an MNIST watermark, we use \texttt{train} split for training data and
\texttt{test} split for the evaluation data. We use non-overlapping data from
\texttt{train} split of the Places365 \citep{z17p} dataset as backgrounds for
the Waterbirds dataset, following \citet{s20d}.

ResNet, VGG, and DeiT were fine-tuned on cloud nodes with 4 A100 GPUs. ConvNeXt
was fine-tuned on cloud nodes with 2 H200 GPUs. For CIFAR and
CelebA, models were trained from scratch on local cluster nodes with 2 H200
GPUs. Finally, for Waterbirds, ResNet-18 models were trained on cloud nodes
with 4 L4 GPUs, and VGG-11 models were trained on cloud nodes with 4 A100 GPUs.
Fine-tuning experiments took around 590 A100-hours and 200 H200-hour. Training
from scratch experiments took around 1245 H200-hours, 765 A100-hours, and 960
L4-hours together.

\begin{table}[htb]
    \centering
    \caption{Number of training epochs}
    \label{tab:epochs}
    \begin{tabular}{c c c c c c}
\toprule
\multicolumn{3}{c}{Fine-tuning} & \multicolumn{3}{c}{Training from scratch} \\
\cmidrule(lr){1-3} \cmidrule(lr){4-6}
CIFAR-10 & Waterbirds & CelebA & CIFAR-10 & Waterbirds & CelebA \\
\midrule
$20$ & $20$ & $2$ & $100$ & $1000$ & $10$ \\
\bottomrule
    \end{tabular}
\end{table}

\begin{table}[htb]
    \centering
    \caption{Hyperparameters for fine-tuning}
    \label{tab:fine-tune-params}
    \begin{tabular}{l l l c}
\toprule
Optimizer & Model & Parameter & Value\\
\midrule
AdamW & DeiT-Ti & \texttt{batch\_size} & $256$\\
& & \texttt{lr} & $1\mathrm{e}{-5} \times \text{\texttt{batch\_size}}^{0.5}$\\
& & \texttt{weight\_decay} & $0.01$\\
& & \texttt{min\_lr} & $1\mathrm{e}{-7} \times \text{\texttt{batch\_size}}^{0.5}$\\
& & Share of warm-up steps & $2\%$\\
\cmidrule(lr){2-4}
& ResNet-18 & \texttt{batch\_size} & $256$\\
& & \texttt{lr} & $1\mathrm{e}{-5} \times \text{\texttt{batch\_size}}^{0.5}$\\
& & \texttt{weight\_decay} & $0.01$\\
& & \texttt{min\_lr} & $1\mathrm{e}{-7} \times \text{\texttt{batch\_size}}^{0.5}$\\
& & Share of warm-up steps & $2\%$\\
\cmidrule(lr){2-4}
& VGG-11 & \texttt{batch\_size} & $256$\\
& & \texttt{lr} & $1\mathrm{e}{-5} \times \text{\texttt{batch\_size}}^{0.5}$\\
& & \texttt{weight\_decay} & $0.01$\\
& & \texttt{min\_lr} & $1\mathrm{e}{-7} \times \text{\texttt{batch\_size}}^{0.5}$\\
& & Share of warm-up steps & $2\%$\\
\cmidrule(lr){2-4}
& ConvNeXt-T & \texttt{batch\_size} & $256$\\
& & \texttt{lr} & $1\mathrm{e}{-5} \times \text{\texttt{batch\_size}}^{0.5}$\\
& & \texttt{weight\_decay} & $0.01$\\
& & \texttt{min\_lr} & $1\mathrm{e}{-7} \times \text{\texttt{batch\_size}}^{0.5}$\\
& & Share of warm-up steps & $2\%$\\
\midrule
SGD & DeiT-Ti & \texttt{batch\_size} & $256$\\
& & \texttt{lr} & $2\mathrm{e}{-5} \times \text{\texttt{batch\_size}}$\\
& & \texttt{weight\_decay} & $0.0001$\\
& & \texttt{momentum} & $0.9$\\
& & \texttt{min\_lr} & $5\mathrm{e}{-7} \times \text{\texttt{batch\_size}}$\\
& & Share of warm-up steps & $2\%$\\
\cmidrule(lr){2-4}
& ResNet-18 & \texttt{batch\_size} & $256$\\
& & \texttt{lr} & $1\mathrm{e}{-4} \times \text{\texttt{batch\_size}}$\\
& & \texttt{weight\_decay} & $0.0001$\\
& & \texttt{momentum} & $0.9$\\
& & \texttt{min\_lr} & $5\mathrm{e}{-7} \times \text{\texttt{batch\_size}}$\\
& & Share of warm-up steps & $2\%$\\
\cmidrule(lr){2-4}
& VGG-11 & \texttt{batch\_size} & $256$\\
& & \texttt{lr} & $1\mathrm{e}{-4} \times \text{\texttt{batch\_size}}$\\
& & \texttt{weight\_decay} & $0.0001$\\
& & \texttt{momentum} & $0.9$\\
& & \texttt{min\_lr} & $5\mathrm{e}{-7} \times \text{\texttt{batch\_size}}$\\
& & Share of warm-up steps & $2\%$\\
\cmidrule(lr){2-4}
& ConvNeXt-T & \texttt{batch\_size} & $256$\\
& & \texttt{lr} & $1\mathrm{e}{-5} \times \text{\texttt{batch\_size}}$\\
& & \texttt{weight\_decay} & $0.0001$\\
& & \texttt{momentum} & $0.9$\\
& & \texttt{min\_lr} & $5\mathrm{e}{-7} \times \text{\texttt{batch\_size}}$\\
& & Share of warm-up steps & $2\%$\\
\bottomrule
    \end{tabular}
\end{table}

\begin{table}[htb]
    \centering
    \caption{Hyperparameters for training from scratch}
    \label{tab:scratch-params}
    \begin{tabular}{l l l c}
\toprule
Optimizer & Model & Parameter & Value\\
\midrule
AdamW & DeiT-Ti & \texttt{batch\_size} & $256$\\
& & \texttt{lr} & $5\mathrm{e}{-5} \times \text{\texttt{batch\_size}}^{0.5}$\\
& & \texttt{weight\_decay} & $0.01$\\
& & \texttt{min\_lr} & $1\mathrm{e}{-7} \times \text{\texttt{batch\_size}}^{0.5}$\\
& & Share of warm-up steps & $5\%$\\
\cmidrule(lr){2-4}
& ResNet-18 & \texttt{batch\_size} & $256$\\
& & \texttt{lr} & $5\mathrm{e}{-3} \times \text{\texttt{batch\_size}}^{0.5}$\\
& & \texttt{weight\_decay} & $0.01$\\
& & \texttt{min\_lr} & $5\mathrm{e}{-5} \times \text{\texttt{batch\_size}}^{0.5}$\\
& & Share of warm-up steps & $5\%$\\
\cmidrule(lr){2-4}
& VGG-18 & \texttt{batch\_size} & $256$\\
& & \texttt{lr} & $5\mathrm{e}{-3} \times \text{\texttt{batch\_size}}^{0.5}$\\
& & \texttt{weight\_decay} & $0.01$\\
& & \texttt{min\_lr} & $5\mathrm{e}{-5} \times \text{\texttt{batch\_size}}^{0.5}$\\
& & Share of warm-up steps & $5\%$\\
\midrule
SGD & DeiT-Ti & \texttt{batch\_size} & $256$\\
& & \texttt{lr} & $2\mathrm{e}{-4} \times \text{\texttt{batch\_size}}$\\
& & \texttt{weight\_decay} & $0.0001$\\
& & \texttt{momentum} & $0.9$\\
& & \texttt{min\_lr} & $5\mathrm{e}{-7} \times \text{\texttt{batch\_size}}$\\
& & Share of warm-up steps & $5\%$\\
\cmidrule(lr){2-4}
& ResNet-18 & \texttt{batch\_size} & $256$\\
& & \texttt{lr} & $5\mathrm{e}{-3} \times \text{\texttt{batch\_size}}$\\
& & \texttt{weight\_decay} & $0.0001$\\
& & \texttt{momentum} & $0.9$\\
& & \texttt{min\_lr} & $2\mathrm{e}{-5} \times \text{\texttt{batch\_size}}$\\
& & Share of warm-up steps & $5\%$\\
\cmidrule(lr){2-4}
& VGG-18 & \texttt{batch\_size} & $256$\\
& & \texttt{lr} & $5\mathrm{e}{-3} \times \text{\texttt{batch\_size}}$\\
& & \texttt{weight\_decay} & $0.0001$\\
& & \texttt{momentum} & $0.9$\\
& & \texttt{min\_lr} & $2\mathrm{e}{-5} \times \text{\texttt{batch\_size}}$\\
& & Share of warm-up steps & $5\%$\\
\bottomrule
    \end{tabular}
\end{table}

\clearpage

\section{MISCELLANEOUS}

\subsection{Licenses of the Used Assets}
\label{sec:licenses}

\paragraph{Datasets}
To the authors' best knowledge, the used datasets have the following licenses
(see \cref{tab:data-lic}).

\begin{table}[htb]
    \centering
    \caption{Licenses of the used datasets}
    \label{tab:data-lic}
    \begin{tabular}{l c}
        \toprule
        Dataset & License (or known restrictions) \\
        \midrule
        MNIST \citep{l98g} & CC BY-SA 3.0 \\
        CIFAR-10 \citep{k09l} & no license specified \\
        CUB-200-2011 \citep{w11c} & non-commercial research and educational restriction \\
        Places365 \citep{z17p} & academic and educational restriction \\
        CelebA \citep{l15d} & Custom non-commercial research license \\
        \bottomrule
    \end{tabular}
\end{table}

\paragraph{Pre-Trained Weights}
To the authors' best knowledge, the used pre-trained models have the following
licenses (see \cref{tab:models-lic}).

\begin{table}[htb]
    \centering
    \caption{Licenses of the used pre-trained models}
    \label{tab:models-lic}
    \begin{tabular}{l c}
        \toprule
        Model & License (or known restrictions) \\
        \midrule
        ResNet-18 & BSD-3 (from the TorchVision library) and non-commercial use
        (from ImageNet) \\
        VGG-11 & BSD-3 (from the TorchVision library) and non-commercial use
        (from ImageNet) \\
        DeiT-Ti & Apache 2.0 (from the original paper) and non-commercial use
        (from ImageNet) \\
        \bottomrule
    \end{tabular}
\end{table}

\end{document}